\documentclass[10pt,twocolumn,letterpaper]{article}

\usepackage{iccv}
\usepackage{times}
\usepackage{epsfig}
\usepackage{graphicx}
\usepackage{amsmath}
\usepackage{amssymb}
\usepackage{multirow}
\usepackage{multicol}
\usepackage{adjustbox}
\usepackage{booktabs}
\usepackage{makecell}
\usepackage{xcolor}
\usepackage{bm}
\usepackage{float}
\usepackage{caption}
\usepackage{latexsym}
\usepackage{subcaption}

\graphicspath{ {./fig/} }

\usepackage[pagebackref=true,breaklinks=true,letterpaper=true,colorlinks,bookmarks=false]{hyperref}

\iccvfinalcopy 

\def\iccvPaperID{1882} 

\ificcvfinal\fi

\begin{document}

\title{Spatial-Temporal Transformer for Dynamic Scene Graph Generation}


\author{Yuren Cong\textsuperscript{1}, Wentong Liao\textsuperscript{1}, Hanno Ackermann\textsuperscript{1}, Bodo Rosenhahn\textsuperscript{1}, Michael Ying Yang\textsuperscript{2}\\
\textsuperscript{1}TNT, Leibniz University Hannover,
\textsuperscript{2}SUG, University of Twente\\


}

\maketitle
\ificcvfinal\thispagestyle{empty}\fi

\begin{abstract}

Dynamic scene graph generation aims at generating a scene graph of the given video. Compared to the task of scene graph generation from images, it is more challenging because of the dynamic relationships between objects and the temporal dependencies between frames allowing for a richer semantic interpretation. In this paper, we propose \textbf{Spatial-temporal Transformer (STTran)}, a neural network that consists of two core modules: (1) a spatial encoder that takes an input frame to extract spatial context and reason about the visual relationships within a frame, and (2) a temporal decoder which takes the output of the spatial encoder as input in order to capture the temporal dependencies between frames and infer the dynamic relationships. Furthermore, STTran is flexible to take varying lengths of videos as input without clipping, which is especially  important for long videos. 
Our method is validated on the benchmark dataset Action Genome (AG).
The experimental results demonstrate the superior performance of our method in terms of dynamic scene graphs. Moreover, a set of ablative studies is conducted and the effect of each proposed module is justified.
Code available at: \url{https://github.com/yrcong/STTran}.
  
\end{abstract}

\section{Introduction} 

A scene graph is a structural representation that summaries objects of interest as nodes and their relationships as edges \cite{johnson2015image,krishna2017visual}.
Recently, scene graphs have been successfully applied in different vision tasks, such as image retrieval \cite{johnson2015image,schuster2015generating}, object detection, semantic segmentation, human-object interaction \cite{gkioxari2018detecting}, image synthesis \cite{johnson2018image,ashual2019specifying}, and high-level vision-language tasks like image captioning \cite{gao2018image,yang2019auto} or visual question answering (VQA) \cite{johnson2017inferring}. It is treated as a promising approach towards holistic scene understanding and a bridge connecting the large gap between vision and natural language domains. 
Therefore, the task of scene graph generation has caught increasing attention in communities.

\begin{figure}[t!]
\begin{center}
   \includegraphics[width=0.99\linewidth]{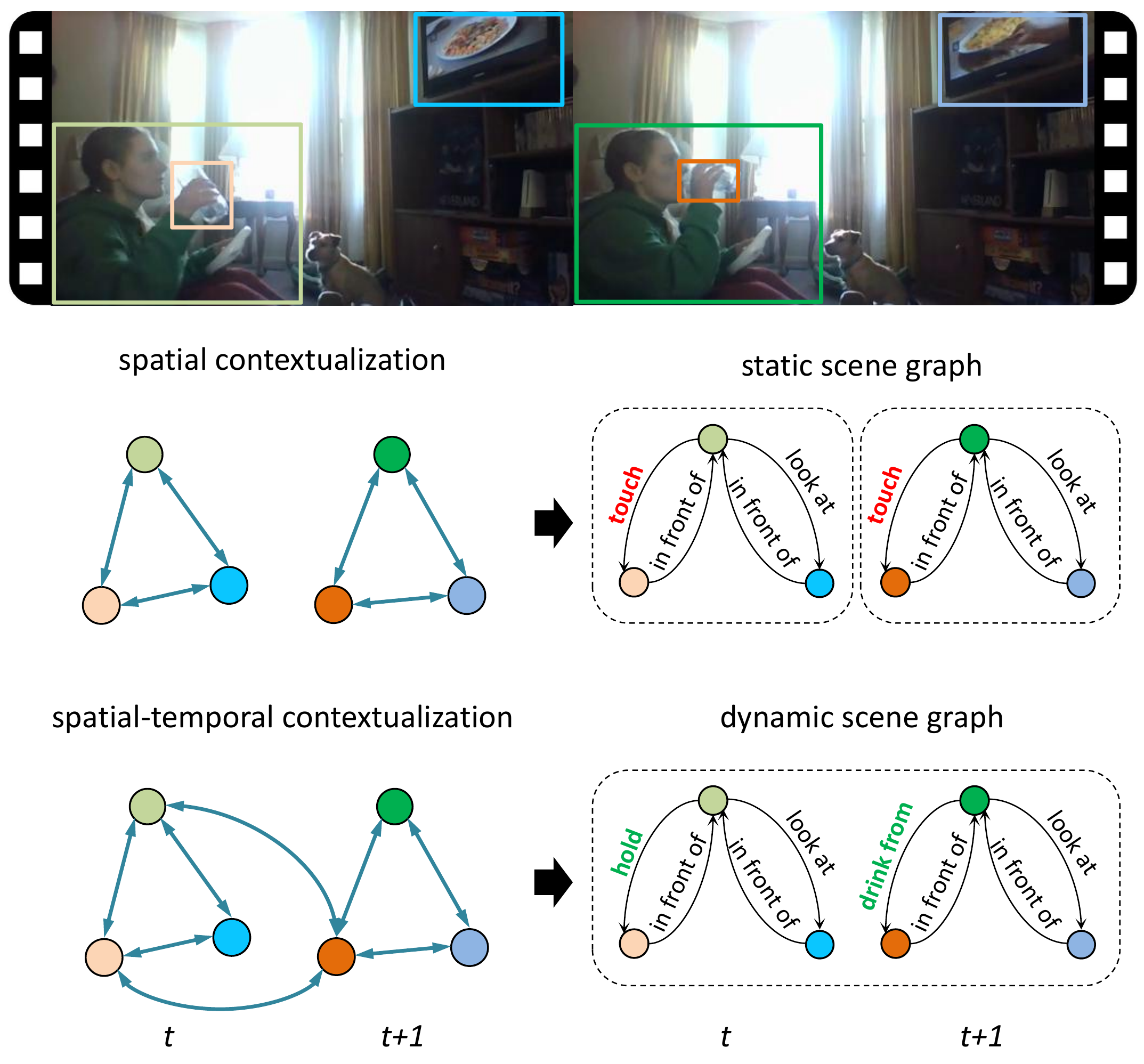}
\end{center}
   \vspace{-5mm}
   \caption{The difference between scene graph generation from image and video. In the video, the person is watching TV and drinking water from the bottle. 
   Dynamic Scene graph generation can utilize both spatial context and temporal dependencies (3rd row) compared with image-based scene graph generation (2nd row). Nodes in different colors denote objects (\texttt{person,bottle,tv})  in the frames.} 
   \vspace{-3mm}
\label{fig:tease}
\end{figure}

While the great progress made in scene graph generation from a single image (static scene graph generation), the task of scene graph generation from a video (dynamic scene graph generation) is new and more challenging.
The most popular approach of static scene graph generation is built upon an object detector that generates object proposals, and then infers their relationship types as well as their object classes.
However, objects are not sure to be consistent in each frame of the video sequence and the relationships between any two objects may vary because of their motions, which is characterized by \emph{dynamic}.
In this case, temporal dependencies play a role, and thus, the static scene graph generation methods are not directly applicable to dynamic scene graph generation, which has been fully discussed in \cite{ji2020action} and verified by the experimental results analyzed in Sec.~\ref{sec:exp}. Fig.~\ref{fig:tease} showcases the difference between scene graph generation from image and video.

Action recognition is an alternative to detect the dynamic relationships between objects. However, actions and activities are typically regarded as monolithic events that occur in videos in action recognition \cite{caba2015activitynet,kay2017kinetics,sigurdsson2016hollywood,li2021pose}.
It has been studied in Cognitive Science and Neuroscience that people perceive an ongoing activity by segmenting them into consistent groups and encoding into a hierarchical part structure \cite{kurby2008segmentation}. 
Let's take the activity "drinking water" as an example, as shown in Fig. \ref{fig:tease}. The person starts this activity by holding the bottle in front of her, and then holds it up and takes water. More complex, the person is looking at the television at the same time. Decomposition of this activity is useful for understanding how it happens and what is going on. Associating with the scene graph, it is possible to predict what will happen: after the person picks up the bottle in front of her, we can predict that the person is likely to  drink water from it. Representing temporal events with structured representations, \ie dynamic scene graph, could lead to more accurate and grounded action understanding.
However, most of the existing methods for action recognition are not able to decompose the activity in this way.

In this paper, we explore how to generate a dynamic scene graph from sequences effectively. The main \textbf{contributions} are summarized as: (1) We propose a novel framework,  Spatial-Temporal Transformer (STTran), which encodes the spatial context within single frames and decodes visual relationship representations with temporal dependencies across frames. (2) Distinct from the majority of related works, multi-label classification is applied in relationship prediction and a new strategy to generate a dynamic scene graph with confident predictions is introduced. (3) With several experiments, we verify that temporal dependencies have a positive effect on relationship prediction and our model improves performance by understanding it. STTran achieves state-of-the-art results on Action Genome \cite{ji2020action}. 

\section{Related Work}

\paragraph{Scene Graph Generation}
Scene graph has first been proposed in \cite{johnson2015image} for image retrieval and  caught increasing attention in Computer Vision community~\cite{lu2016visual,yang*17:graph,li2017scene,dai2017detecting,liao2019natural,tang2019learning,wang2019exploring,yang2019auto,yu2017visual,liao*21:graph}.
It is a graph-based representation describing interactions between objects in the image. 
Nodes in the scene graph indicate the objects while edges denote the relationships.  
The applications include image retrieval \cite{schuster2015generating}, image captioning \cite{anderson2016spice,ren2015faster}, VQA \cite{tang2019learning,johnson2017inferring} and image generation \cite{johnson2018image, he2021context}. In order to generate high-quality scene graphs from images, a series of works explore different directions such as utilizing spatial context \cite{yang*17:graph,zellers2018neural,lin2020gps}, graph structure \cite{yang2018graph,xu2017scene,li2018factorizable}, optimization \cite{cong2020nodis}, reinforcement learning \cite{liang2017deep,tang2019learning}, semi-supervised training \cite{chen2019scene} or a contrastive loss \cite{zhang2019graphical}. These works have achieved excellent results on image datasets \cite{krishna2017visual,lu2016visual,kuznetsova2020open}. 
Although it is universal for multiple relationships to co-occur between a subject-object pair in the real world, the majority of previous works defaults to edge prediction as single-label classification. 
Despite the progress made in this field,  all these methods are designed for static images. 
In order to extend the gain brought by scene graphs in images to video, Ji \textsl{et al.} \cite{ji2020action} collect a large dataset of dynamic scene graphs by decomposing activities in videos and improve state of the art results for video action recognition with dynamic scene graph. 

\vspace{-4mm}
\paragraph{Transformer for Computer Vision}
The vanilla Transformer architecture was proposed by Vaswani \etal \cite{vaswani2017attention} for neural machine translation. Many transformer variants are developed and have achieved great performance in language modeling tasks, especially the large-scale pre-trained language models, like GPT \cite{radford2019language} and BERT \cite{devlin2018bert}. 
Then, Transformers have also been widely and successfully applied in many vision-language tasks, such as image captioning \cite{xu2015show,he2020image}, VQA \cite{anderson2018bottom,yang2020bert}. To further bridge the vision and language domains, different Bert-like large-scale pre-trained models are also developed
, like Caption-Based Image Retrieval and Visual Commonsense Reasoning (VCR)  \cite{lu2019vilbert,li2020does,Su2020VL-BERT}.
Most recently, Transformers are attracting increasing attention in the vision community. DETR is introduced by Carion \etal \cite{carion2020end} for object detection and panoptic segmentation. 
Moreover, Transformers are explored to learn vision features from the given image instead of the traditional CNN backbones and achieve promising performance \cite{dosovitskiy2020image,touvron2020training}.
The core mechanism of Transformer is its self-attention building block which is able to make predictions by selectively attending to the input points (each point can be a word representation of a sentence or a local feature from an image), so that context is captured between different input points and the representation of each point is refined. 
Nonetheless, the above methods focus on learning spatial context with a transformer from a single image while temporal dependencies play a role in video understanding. 
Action Transformer is proposed by Girdhar \etal \cite{girdhar2019video} that utilizes transformer to refine the spatio-temporal representations, which are learned by I3D model \cite{carreira2017quo} and then pooled from the RoI given by a RPN network \cite{ren2015faster}, for recognizing human actions in video clips. In fact, the transformer module is still used to learn spatial context.
VisTR is introduced in \cite{wang2020end} for video segmentation. The features of each frame that are extracted by a CNN backbone are fed to a transformer encoder to learn the temporal information of a video sequence.

\vspace{-4mm}
\paragraph{Spatial-Temporal Networks}
Spatial-temporal information is the key to access video understanding \cite{lin2019tsm, kluger2020temporally, hornakova2020lifted} and has been long and well studied. To date, the most popular approaches are RNN/LSTM-based \cite{hochreiter1997long} or 3D ConvNets-based \cite{ji20123d,tran2015learning} structures. The former takes features from each frame sequentially and learns the temporal information \cite{srivastava2015unsupervised,donahue2015long}. The latter extends the traditional 2D convolution (height and width dimension) to time dimension for sequential inputs.
Simonyan \etal \cite{simonyan2014two} introduce a two-stream CNN structure that spatial and temporal information is learned on different streams respectively. Residual connections are inserted between the two information streams to allow information fusion.
Then, the 2D convolution in the two-stream structure is inflated into its counterpart 3D convolution, dubbed I3D model \cite{carreira2017quo}. 
Non-local Neural Networks \cite{wang2018non} introduce another kind of generic self-attention mechanism, non-local operation. 
It computes relatedness between different locations in the input signal and refines the inputs by weighted sum of different inputs based on the relatedness. Their method is easy to be applied in video input by extending the non-local operation along the time dimension.
However, these works are applied for activity recognition and are not able to decompose the activity into consistent groups.
In this work, we do not only utilize transformer to learn spatial context between objects within a frame, but also the temporal dependencies between frames to infer the dynamic relationships varying along the time axis.

\section{Method}
\begin{figure*}[ht!]
\centering
\includegraphics[width=0.99\linewidth]{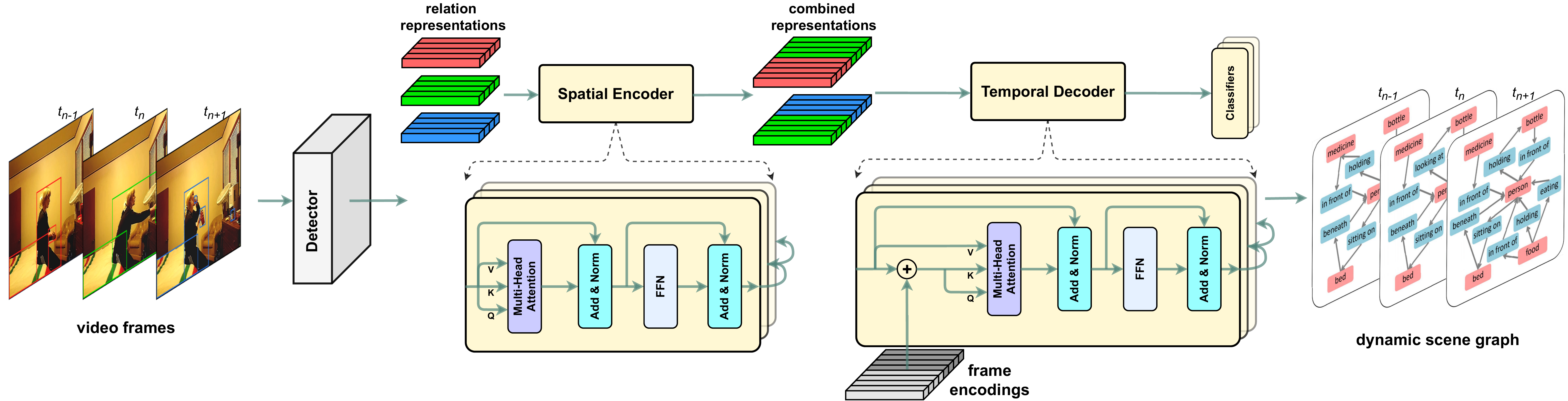}
\vspace{-1mm}
\caption{Overview of our method: the object detection backbone proposes object regions in RGB video frames and the relationship feature vectors are pre-processed (Sec.~\ref{Sec:representation}). The encoder of the proposed Spatial-Temporal Transformer (Sec.~\ref{Sec:sttran}) first extracts the spatial context within single frames. The relation representations refined by encoder stacks from different frames are combined and added to learned frame encodings. The decoder layers capture temporal dependencies and relationships are predicted with linear classifiers for different relation type (such as $attention$, $spatial$, $contact$). $\oplus$ indicates element-wise addition while FFN stands for feed-forward network.}
\vspace{-5mm}
\label{fig:model}
\end{figure*}

A dynamic scene graph $G_{dyn}(\mathcal{V}_{t},\mathcal{E}_{t})$ can be modeled as a static scene graph $G_{stat}(\mathcal{V},\mathcal{E})$ with an extra index $t$ representing the relations over time as an extra temporal axis.
Inspired by the transformer characteristics: (1) the architecture is permutation-invariant, and (2) the sequence is compatible with positional encoding, we introduce a novel model, Spatial-Temporal Transformer (\textbf{STTran}), in order to utilize the spatial-temporal context along videos (see Fig.~\ref{fig:model}).

\subsection{Transformer}
\label{Sec:transformer}
First, we take a brief review on the transformer structure.
The transformer is proposed by Vaswani \etal \cite{vaswani2017attention} and consists of a stack of multi-head dot-product attention based transformer refining layers. 
In each layer, the input $\bm{X} \in \mathbb{R}^{N\times D}$ that has $N$ entries of $D$ dimensions, is transformed into queries ($\bm{Q} = \bm{X}\bm{W}_Q$, $\bm{W}_Q \in \mathbb{R}^{D\times D_q}$), keys ($\bm{K}= \bm{X}\bm{W}_K$, $\bm{W}_K \in \mathbb{R}^{D\times D_k}$) and values ($\bm{V} = \bm{X}\bm{W}_V$, $\bm{W}_V \in \mathbb{R}^{D\times D_v}$) though linear transformations. Note that $D_q$, $D_k$ and $D_v$ are the same in the implementation normally.
Each entry is refined with other entries through dot-product attention defined by:
\begin{equation}\label{eq:attention}
\centering
\begin{aligned}
  Attention(\bm{Q},\bm{K},\bm{V}) = Softmax\left(\frac{\bm{Q}\bm{K}^T}{\sqrt{D_k}}\right) \bm{V},
\end{aligned}
\end{equation}
To improve the performance of the attention layer, multi-head attention is applied which is defined as :
\begin{equation}
\centering
\begin{aligned}
& MultiHead(\bm{Q},\bm{K},\bm{V}) = Concat(h_1,\dots,h_h)\bm{W}_O,\\
& h_i = Attention(\bm{X}\bm{W}_{Q_i},\bm{X}\bm{W}_{K_i},\bm{X}\bm{W}_{V_i}).
\end{aligned}
\label{eq:multiheawd_1}
\end{equation}
A complete self-attention layer contains the above self-attention module followed by a normalization layer with residual connection and a feed-forward layer, which is also followed by a normalization layer with residual connection. For simplicity, we denote such a self-attention layer as $Att(.)$.
In this work, we design a Spatio-Temporal Transformer based on $Att(.)$ to explore the spatial context, which works on a single frame, and temporal dependencies that work on sequence, respectively.

\subsection{Relationship Representation}
\label{Sec:representation}
We employ Faster R-CNN \cite{ren2015faster} as our backbone. 
For the frame $I_t$ at time step $t$ in a given video with $T$ frames $V=[I_1,I_2,\dots,I_T]$, the detector provides visual features $\lbrace\bm{v}^1_t,\dots,\bm{v}^{N(t)}_t\rbrace_\in\mathbb{R}^{2048}$, bounding boxes $\lbrace\bm{b}^1_t,\dots,\bm{b}^{N(t)}_t\rbrace$ and object category distribution $\lbrace\bm{d}^1_t,\dots,\bm{d}^{N(t)}\rbrace$ of object proposals where $N(t)$ indicates the number of object proposals in the frame. Between the $N(t)$ object proposals there is a set of relationships $R_t=\lbrace r^1_t,r^2_t,\dots,r^{K(t)}_t\rbrace$. The representation vector $\bm{x}^k_t$ of the relation $r^k_t$ between the $i$-th and $j$-th object proposals contains visual appearances, 
spatial information and semantic embeddings, which can be formulated as: 
    \begin{equation}
        \bm{x}^k_t = \left< \bm{W}_{s}\bm{v}^i_t,\bm{W}_{o}\bm{v}^j_t,\bm{W}_{u}\varphi(\bm{u}^{ij}_t\oplus f_{box}(\bm{b}^i_t,\bm{b}^j_t))),\bm{s}^i_t,\bm{s}^j_t \right>
    \label{Eq:rel_features}
    \end{equation}
where $\left<,\right>$ is concatenation operation, $\varphi$ is flattening operation and $\oplus$ is element-wise addition. 
$\bm{W}_s$, $\bm{W}_o\in\mathbb{R}^{2048\times512}$ and $\bm{W}_u\in\mathbb{R}^{12544\times512}$ represent the linear matrices for dimension compression. 
$\bm{u}^{ij}_t\in\mathbb{R}^{256\times7\times7}$ indicates the feature map of the union box computed by RoIAlign \cite{he2017mask} while $f_{box}$ is the function transforming the bounding boxes of subject and object to an entire feature with the same shape as $\bm{u}^{ij}_t$. The semantic embedding vectors $\bm{s}^i_t$, $\bm{s}^j_t\in\mathbb{R}^{200}$ are determined by the object categories of subject and object. The relationship representations exchange spatial and temporal information in Spatial-Temporal Transformer.

\subsection{Spatio-Temporal Transformer}
\label{Sec:sttran}

The Spatio-Temporal Transformer maintains the original encoder-decoder architecture \cite{vaswani2017attention}. The difference is, the encoder and decoder are delegated the more concrete tasks.

\vspace{-4mm}
\paragraph{Spatial Encoder}
concentrates on the spatial context within a frame whose input is a single $\bm{X}_t=\lbrace \bm{x}^1_t,\bm{x}^2_t,\dots,\bm{x}^{K(t)}_t\rbrace$. 
The queries $\bm{Q}$, keys $\bm{K}$ and values $\bm{V}$ share the same input and the output of the $n$-th encoder layer is presented as:
    \begin{equation}
         \bm{X}^{(n)}_t = Att_{enc.}(\bm{Q}=\bm{K}=\bm{V}=\bm{X}^{(n-1)}_t)
    \label{Eq:enc_formula}
    \end{equation}
The encoder consists of N identical $Att_{enc.}$ layers that are stacked sequentially. The input of the $(n)$-th layer is the output of the $(n-1)$-th layer. For simplicity, we remove the superscript $n$ in the following discussion.
Unlike the majority of transformer methods, no additional position encoding is integrated into the inputs since the relationships within a frame are intuitively parallel. Having said that, the spatial information hiding in the relation representations (see Eq.~\ref{Eq:rel_features}) plays a crucial role in the self-attention mechanism. The final output of the encoder stacks is sent to the Temporal Decoder.

\vspace{-4mm}
\paragraph{Frame Encoding}
is introduced for the temporal decoder. Without convolution and recurrence, the knowledge of sequence order such as positional encoding must be embedded in the input for the transformer. In contrast to the word position in \cite{vaswani2017attention} or the pixel position in \cite{carion2020end}, we customize the frame encodings to inject the temporal position in the relationship representations. The frame encodings  $\bm{E}_{f}$ are constructed with learned embedding parameters, since the amount of the embedding vectors depending on the window size $\eta$ in the Temporal Decoder is fixed and relative short:
        $ \bm{E}_{f} = [\bm{e}_{1},\dots,\bm{e}_{\eta}], $
where $\bm{e}_{1},\dots,\bm{e}_{\eta}\in\mathbb{R}^{1936}$ are the learned vectors with the same length as $\bm{x}^k_t$.

The widely used sinusoidal encoding method is also analyzed (see Table \ref{tab:ablation}). We adopt the learned encoding method because of its overall better performance. The window size $\eta$ is fixed and therefore the video length does not affect the length of frame encodings.

\vspace{-4mm}
\paragraph{Temporal Decoder} captures the temporal dependencies between frames. 
Not only the amount of calculation required and the memory consumption increase greatly, but also useful information is easily overwhelmed by a large number of irrelevant representations. 
In this work, we adopt a sliding window to batch the frames so that the message is passed between the adjacent frames in order to avoid interference with distant frames.

Different from \cite{vaswani2017attention}, the self-attention layer of our temporal decoder is identical to the spatial encoder $Att_{enc.}()$, \ie the masked multi-head self-attention layers are removed. A sliding window of size $\eta$ runs over the sequence of spatial contextualized representations $[\bm{X}_1,\dots,\bm{X}_T]$ and the $i$-th generated input batch is presented as: 
    \begin{equation}
         \bm{Z}_{i} = [\bm{X}_{i},\dots,\bm{X}_{i+\eta-1}], i\in\lbrace1,\dots,T-\eta+1 \rbrace
    \label{Eq:batch_input}
    \end{equation}
where the window size $\eta\leq T$ and $T$ is the video length. The decoder consists of $N$ stacked identical self-attention layer $Att_{dec}()$ similar as the encoder structure. Considering the first layer:
\begin{equation}
\centering
\begin{aligned}
& \bm{Q} = \bm{K} = \bm{Z}_i + \bm{E}_{f},\\
& \bm{V} = \bm{Z}_i, \\
&\hat{\bm{Z}}_i = Att_{dec.}(\bm{Q},\bm{K},\bm{V}).
\end{aligned}
\label{eq:multiheawd}
\end{equation}
Regarding the first line in Eq.~\ref{eq:multiheawd}, same encoding is added to the relation representations in the same frame as queries and keys. The output from the last decoder layer is adopted for final prediction. 
Because of the sliding widow, the relationships in a frame have various representation in different batches.
In this work, we choose the earliest representation appearing in the windows.

\subsection{Loss Function}
We employ multiple linear transformations to infer different kinds of relationships (such as attention, spatial, contacting) with the refined representations. In reality, the same type of relationship between two objects is not unique in semantics, such as synonymous actions \texttt{$<$person-holding-broom$>$} and \texttt{$<$person-touching-broom$>$}. Thereby, we introduce the multi-label margin loss function for predicate classification as follows:
    \begin{equation}
        L_{p}(r,\mathcal{P}^{+},\mathcal{P}^{-}) = \sum_{p\in\mathcal{P}^{+}} \sum_{q\in\mathcal{P}^{-}}max(0,1-\phi(r,p)+\phi(r,q))
    \label{Eq:pred_loss1}
    \end{equation}
For a subject-object pair $r$, $\mathcal{P}^{+}$ are the annotated predicates while 
$\mathcal{P}^{-}$ is the set of the predicates not in the annotation. $\phi(r,p)$ indicates the computed confidence score of the $p$-th predicate. 

During training, the object distribution is computed by two fully-connected layers with a ReLU activation and a batch normalization in between. The standard cross entropy loss $L_o$ is utilized. The total objective is formulated as:
    \begin{equation}
        L_{total} = L_p+L_o
    \label{Eq:pred_loss2}
    \end{equation}

\subsection{Graph Generation Strategies}
There are two typical strategies to generate a scene graph with the inferred relation distribution in previous works: (a) \textbf{With Constraint} only allows each subject-object pair to have at most one predicate while (b) \textbf{No Constraint} allows a subject-object pair to have multiple edges in the output graph with multiple guesses. \textbf{With Constraint} is more rigorous and indicates the ability of models to predict the most important relationships, but it is incompetent for the multi-label task. Although \textbf{No Constraint} can reflect the ability of multi-label prediction, tolerant multiple guesses cause wrong information in the generated scene graph.

In order to make the generated scene graph closer to ground truth, we propose a new strategy named \textbf{Semi Constraint} allowing that a subject-object pair has multiple predicates such as \texttt{$<$person-holding-food$>$} and \texttt{$<$person-eating-food$>$}. The predicate is regarded as positive iff the corresponding relation confidence is higher than the threshold. 

At test time, the score of each relationship triplet \texttt{$<$subject-predicate-object$>$} is computed as:
    \begin{equation}
        s_{rel} = s_{sub}\cdot s_{p}\cdot s_{obj},
    \label{Eq:pred_loss3}
    \end{equation}
where  $s_{sub}$,$s_{p}$,$s_{obj}$ are the confidence score of subject, predicate and object respectively.

\section{Experiments}
\label{sec:exp}

\subsection{Dataset and Evaluation Metrics}
\label{Sec:dataset}

\paragraph{Dataset}
We train and validate our model on the Action Genome (AG) dataset \cite{ji2020action} which provides frame-level scene graph labels and is built upon the Charades dataset \cite{sigurdsson2016hollywood}. $476, 229$ bounding boxes of 35 object classes (without $person$) and $1,715,568$ instances of 25 relationship classes are annotated for $234, 253$ frames. These 25 relationships are subdivided into three different types: (1) \textsl{attention} relationships denoting whether a person is looking at an object, (2)  \textsl{spatial} relationships and (3) \textsl{contact} relationships which indicate the different ways the object is contacted. In AG, $135,484$ subject-object pairs are labeled with multiple spatial relationships (\eg\texttt{$<$door-in front of-person$>$} and \texttt{$<$door-on the side of-person$>$}) or contact relationships (\eg\texttt{$<$person-eating-food$>$} and \texttt{$<$person-holding-food$>$}). 

\vspace{-2mm}
\paragraph{Evaluation Metrics}
We follow three standard tasks from image-based scene graph generation \cite{lu2016visual} for evaluation : (1) predicate classification (PREDCLS): given ground truth labels and bounding boxes of objects, predict predicate labels of object pairs. (2) scene graph classification (SGCLS): classify the ground truth bounding boxes and predict relationship labels. (3) Scene graph detection (SGDET): detect the objects and predict relationship labels of object pairs. The object detection is regarded as successful if the predicted box overlaps with the ground-truth box at least 0.5 IoU. All tasks are evaluated with the widely used $Recall@K$ metrics ($K=[10,20,50]$) following \textbf{With Constraint}, \textbf{Semi Constraint} and \textbf{No Constraint}. The threshold of confidence in the relationship is set to $0.9$ in \textbf{Semi Constraint} for all experiments if no special instruction.

\subsection{Technical Details}
\label{Sec:implementation_details}
In this work, FasterRCNN \cite{ren2015faster} based on ResNet101 \cite{he2016deep} is adopted as object detection backbone. We first train the detector on the training set of Action Genome \cite{ji2020action} and get 24.6 mAP at 0.5 IoU with COCO metrics. The detector is applied to all baselines for fair comparison. The parameters of the object detector including RPN are fixed when training scene graph generation models. Per-class non-maximal suppression at 0.4 IoU is applied to reduce region proposals provided by RPN.

We use an AdamW \cite{loshchilov2017decoupled} optimizer with initial learning rate $1e^{-5}$ and batch size 1 to train our model. Moreover, gradient clipping is applied with a maximal norm of 5. For all experiments on Action Genome, we set the window size $\eta =2$ and $stride = 1$ for our STTran. The spatial encoder contains 1 layer while the temporal decoder contains 3 iterative layers. The self-attention module in both encoder and decoder has 8 heads with $d_{model}=1936$ and $dropout=0.1$. 
The $\texttt{1936-d}$ input is projected to $\texttt{2048-d}$ by the feed-forward network, then projected to $\texttt{1936-d}$ again after ReLU activation.

\begin{table*}[!htbp]
\centering
\begin{adjustbox}{width=1\textwidth}
\begin{tabular}{ccccccccccccccccccc}
 \hline \hline
 \multirow{3}*{Method} &  \multicolumn{9}{c}{With Constraint} &\multicolumn{9}{c}{No Constraint} \cr
    \cmidrule(lr){2-10} \cmidrule(lr){11-19}  
  & \multicolumn{3}{c}{PredCLS} & \multicolumn{3}{c}{SGCLS} & \multicolumn{3}{c}{SGDET} & \multicolumn{3}{c}{PredCLS} & \multicolumn{3}{c}{SGCLS} & \multicolumn{3}{c}{SGDET} \cr
    \cmidrule(lr){2-4} \cmidrule(lr){5-7} \cmidrule(lr){8-10} \cmidrule(lr){11-13} \cmidrule(lr){14-16} \cmidrule(lr){17-19} 
  &R@10 &R@20 &R@50 &R@10 &R@20 &R@50 &R@10 &R@20 &R@50 &R@10  &R@20 &R@50 &R@10 &R@20 &R@50 &R@10 &R@20 &R@50\\
  \hline \hline
  VRD\cite{lu2016visual}& 51.7& 54.7& 54.7& 32.4& 33.3& 33.3& 19.2& 24.5& 26.0& 59.6& 78.5& 99.2& 39.2& 49.8& 52.6& 19.1& 28.8& 40.5\\
  Motif Freq\cite{zellers2018neural}& 62.4& 65.1& 65.1& 40.8& 41.9& 41.9& 23.7& 31.4& 33.3& 73.4& 92.4&  \textbf{99.6}& 50.4& 60.6& 64.2& 22.8& 34.3& 46.4\\
  MSDN\cite{li2017scene} & 65.5& 68.5& 68.5& 43.9& 45.1& 45.1& 24.1& 32.4& 34.5& 74.9& 92.7& 99.0& 51.2& 61.8& 65.0& 23.1& 34.7& 46.5\\
  VCTREE\cite{tang2019learning}& 66.0& 69.3& 69.3& 44.1& 45.3& 45.3& 24.4& 32.6& 34.7& 75.5& 92.9& 99.3& 52.4& 62.0& 65.1& 23.9& 35.3& 46.8\\
  RelDN\cite{zhang2019graphical} & 66.3& 69.5& 69.5& 44.3& 45.4& 45.4& 24.5& 32.8& 34.9& 75.7& 93.0& 99.0& 52.9& 62.4& 65.1& 24.1& 35.4& 46.8\\
  GPS-Net\cite{lin2020gps}& 66.8& 69.9& 69.9& 45.3& 46.5& 46.5& 24.7& 33.1& 35.1& 76.0& 93.6& 99.5& 53.6& 63.3& 66.0& 24.4& 35.7 &47.3\\
  STTran& \textbf{68.6}& \textbf{71.8}& \textbf{71.8}& \textbf{46.4}& \textbf{47.5}& \textbf{47.5}& \textbf{25.2} & \textbf{34.1}& \textbf{37.0}& \textbf{77.9}& \textbf{94.2}& 99.1& \textbf{54.0}& \textbf{63.7}& \textbf{66.4}& \textbf{24.6}& \textbf{36.2} &\textbf{48.8}\\
\hline \hline
\end{tabular}
\end{adjustbox}
\caption{Comparison with state-of-the-art image-based scene graph generation methods on Action Genome \cite{ji2020action}.The same object detector is used in all baselines for fair comparison. STTran has the best performance in all metrics. Note that the evaluation results of baselines are different from \cite{ji2020action} since we adopted a more reasonable relationship output method, more details are provided in the supplementary material.} 
\label{tab:quantitative_result}
\end{table*}

\begin{table}[!htbp]
\centering
\begin{adjustbox}{max width=0.48\textwidth}
\begin{tabular}{cccccccccc}
 \hline \hline
 \multirow{3}*{Method} &  \multicolumn{9}{c}{Semi Constraint} \cr
    \cmidrule(lr){2-10} 
  & \multicolumn{3}{c}{PredCLS} & \multicolumn{3}{c}{SGCLS} & \multicolumn{3}{c}{SGDET} \cr
    \cmidrule(lr){2-4} \cmidrule(lr){5-7} \cmidrule(lr){8-10} 
  &R@10 &R@20 &R@50 &R@10 &R@20 &R@50 &R@10 &R@20 &R@50 \\
  \hline \hline
  VRD\cite{lu2016visual}& 55.5& 64.9& 65.2& 36.2& 39.7& 40.1& 19.0& 27.1& 32.4\\
  Motif Freq\cite{zellers2018neural}& 65.7& 74.1& 74.5& 45.5& 49.3& 49.5& 22.9& 33.7& 39.0\\
  MSDN\cite{li2017scene} & 69.6& 78.9& 79.9& 48.3& 54.1& 54.5& 23.2& 34.2& 41.5\\
  VCTREE\cite{tang2019learning}& 70.1& 78.2& 79.6& 49.0& 53.7& 54.0& 23.7& 34.8& 40.4\\
  RelDN\cite{zhang2019graphical} & 70.7& 78.8& 80.3& 49.4& 53.9& 54.1& 24.1& 35.0& 40.7\\
  GPS-Net\cite{lin2020gps}& 71.3& 81.2& 82.0& 50.2& 55.0& 55.2& 24.5& 35.3& 41.9\\
  STTran& \textbf{73.2}& \textbf{83.1}& \textbf{84.0}& \textbf{51.2}& \textbf{56.5}& \textbf{56.8}& \textbf{24.6} & \textbf{35.9} &\textbf{44.0}\\
\hline \hline
\end{tabular}
\end{adjustbox}
  \caption{Evaluation results of \textbf{Semi Constraint} which indicates the relationship between object pair is regarded as positive if the confidence score is higher than the threshold.}
  \vspace{-1mm}
  \label{tab:semi_result}
\end{table}

\subsection{Quantitative Results and Comparison}
\label{Sec:comparison}
Table \ref{tab:quantitative_result} shows that our model outperforms state-of-the-art image-based methods in all metrics following \textbf{With Constraint}, \textbf{Semi Constraint} and \textbf{No Constraint}. For the fair comparison, all methods share the identical object detector which provides feature maps and region proposals of the same quality.

The bold numbers denote the best result in any column. With the help of temporal dependencies our model improves state-of-the-art (GPS-Net \cite{lin2020gps}) $1.9\%$ on PredCLS-$R@20$, $1.0\%$ on SGCLS-$R@20$ and $1.0\%$ on  SGDET-$R@20$  for the strategy \textbf{With Constraint}, which shows that STTran performs better than image-based baselines in predicting the most important relationships between an object pair. 
Our model also has excellent performance (see Table \ref{tab:semi_result}): $1.9\%$ on PredCLS-$R@20$, $1.5\%$ on SGCLS-$R@20$ and $0.6\%$ improvement on SGDET-$R@20$ for \textbf{Semi Constraint} that allows multiple relationships between a subject-object pair. For \textbf{No Constraint}, STTran outperforms other methods in all settings except PredCLS-$R@50$. Due to the small number of object pairs and the large number (50) of chances to guess, the results in this column are unstable and unconvincing. Motif Freq \cite{zellers2018neural} which is very dependent on statistics achieves the highest score. However, the results become reliable with the less prediction number $K=[10,20]$.

Note that there is no difference between PredCLS-$R@20$ and PredCLS-$R@50$ for \textbf{With Constraint} because of a limited number of object pairs and edge restriction. This also happens on SGCLS. Compared with PredCLS or SGCLS, the gap of SGDET between STTran and other methods is narrowed since the increased false object proposals cause interference, especially for \textbf{Semi Constraint} and \textbf{No Constraint} using small $K$. Furthermore, the reproduced results of some methods are different from \cite{ji2020action} since a more reasonable relationship output method is adopted and the object detectors are different. 

\begin{figure}[http]
\centering
\includegraphics[width=0.99\linewidth]{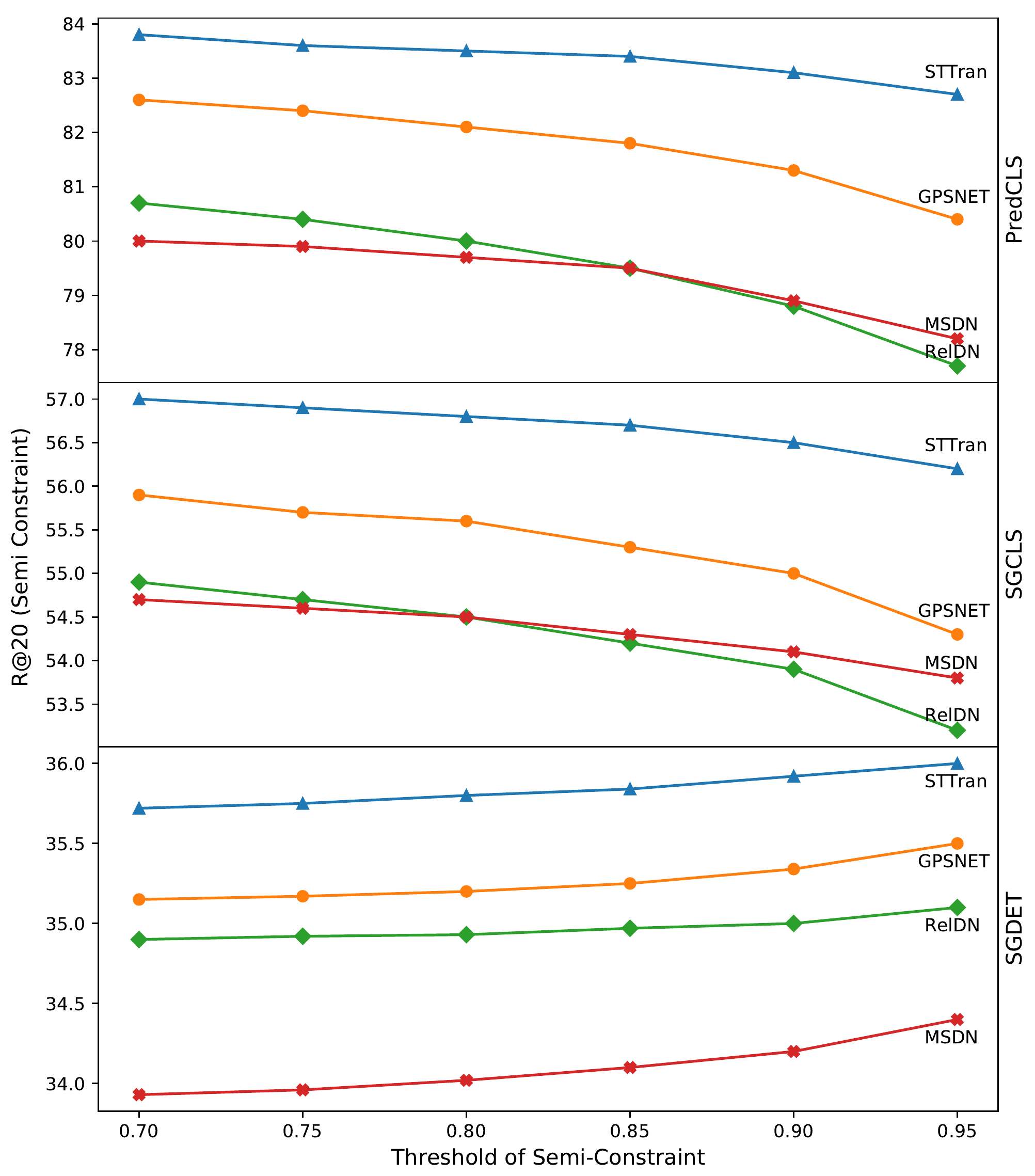}
\caption{$R@20$-Threshold Curves in three standard tasks (PredCLS$/$SGCLS$/$SGDET) for \textbf{Semi Constraint}.} 
\label{fig:semi_threshold}
\vspace{-4mm}
\end{figure}

In \textbf{Semi Constraint}, the threshold of confidence in the relationship is set to a fixed number ($0.9$) in the experiments.
In order to study the impact of such threshold in \textbf{Semi Constraint} on $Recall@K$, the $R@20$-Threshold curves of \cite{li2017scene, zhang2019graphical, lin2020gps} and STTran are shown in Fig.~\ref{fig:semi_threshold}. STTran consistently outperforms all three models at all threshold levels from $0.7$ to $0.95$. The high threshold suppresses the $R@20$ values except in SGDET since there are more pair proposals.

\subsection{Temporal Dependency Analysis}
\label{Sec:temporal_analysis}
Compared to the previous image-based scene graph generation, a dynamic scene graph has  additional temporal dependencies that can be utilized. We discuss whether temporal dependencies can improve the relationship inference and validate that our proposed method utilize temporal dependencies. In this subsection, we measure PredCLS-$R@20$ (\textbf{With Constraint}) as the performance indicator that shows the ability of single relationship classification strictly.

  \vspace{-4mm}
\paragraph{Is temporal dependence easy to use?}
Spatial context plays a relevant role in scene graph generation as validated by several image-based methods \cite{zellers2018neural,lin2020gps}. To explore the effectiveness of temporal dependencies, we graft the widely-used recurrent network, LSTM onto the baselines in Table \ref{tab:baseline_lstm} as follows. Before forwarding the feature vectors into the final classifiers, the entire vectors representing relationships in the video are organized as a sequence and processed by LSTM.

Table \ref{tab:baseline_lstm} shows all baselines can gain more or less from the temporal dependencies. For Motif Freq \cite{zellers2018neural}, PredCLS-$R@20$ increases from 65.1\% to 65.2\% slightly probably due to the relatively simple feature representation. Meanwhile, the score of GPS-Net \cite{lin2020gps} is improved from 69.9\% to 70.4\% significantly. The experiment  shows that temporal dependencies are helpful for scene graph generation. However, the previous methods were designed for static images. This is why we propose Spatial-Temporal Transformer (STTran) to make better use of temporal dependencies. 

\begin{table}[htbp]
\centering
\begin{adjustbox}{max width=0.25\textwidth}
\begin{tabular}{ccc}
 \hline \hline
 \multirow{2}*{Method}  &\multicolumn{2}{c}{PredCLS-R@20}\cr
  \cmidrule(lr){2-3} 
 & original  & +LSTM \\
  \hline \hline
Motif Freq\cite{zellers2018neural}& 65.1& 65.2 \\
MSDN\cite{li2017scene} & 68.5 & 68.8  \\
RelDN\cite{zhang2019graphical} & 69.5 & 69.7  \\
GPS-Net\cite{lin2020gps} & 69.9 & 70.4  \\
\hline \hline
\end{tabular}
\end{adjustbox}
\caption{We integrate LSTMs to process the relationship features before forwarding them into the classifier into some representative baselines. All baselines are improved with temporal dependencies but worse than our STTran.} 
  \vspace{-4mm}
\label{tab:baseline_lstm}
\end{table}

\begin{figure}[http]
\centering
\begin{subfigure}[b]{0.48\linewidth}
  \centering
  \includegraphics[width=1\textwidth]{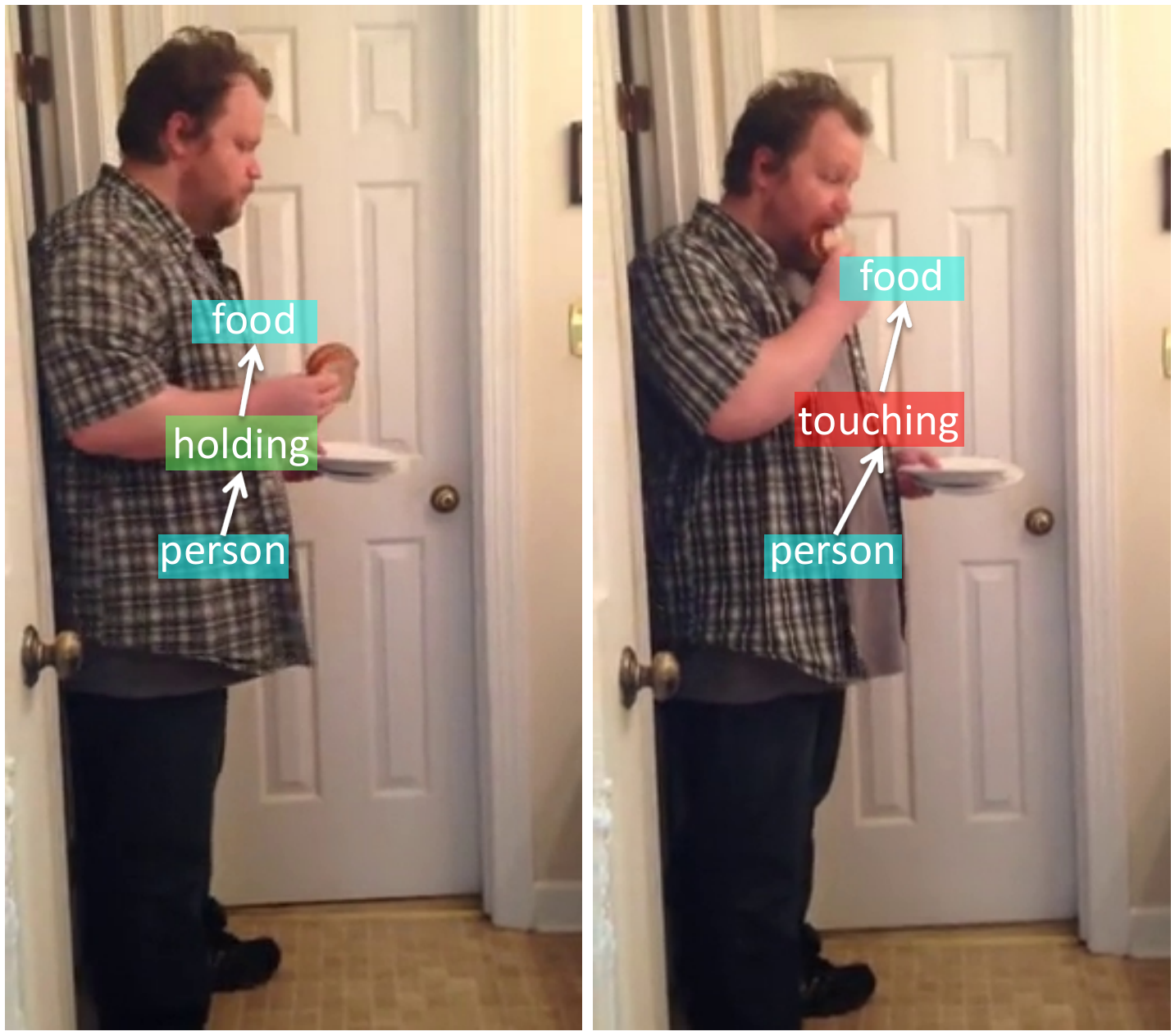}  
  \caption{spatial encoder only}
\end{subfigure}
\begin{subfigure}[b]{0.48\linewidth}
  \centering
  \includegraphics[width=1\textwidth]{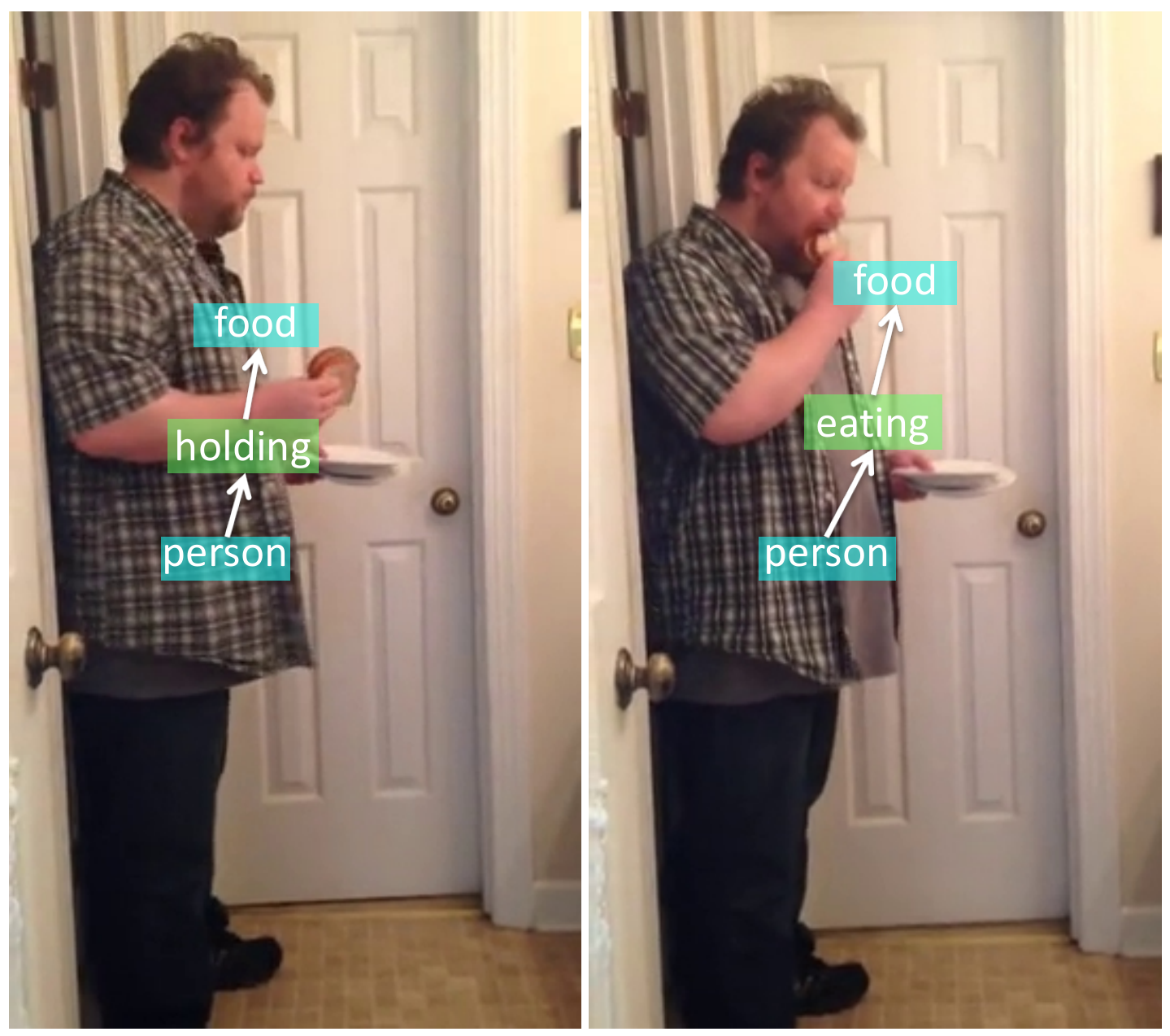}  
  \caption{complete STTran}
\end{subfigure}
\caption{Two relationship instances respectively generated by the spatial encoder and STTran. (a) Spatial encoder predicts the wrong relationship only with the spatial context in the second frame while (b) STTran can infer more accurate results with the help of temporal dependencies.}
  \vspace{-4mm}
\label{fig:ablation_instance}
\end{figure}

\begin{figure*}[t!]
\centering
\includegraphics[width=0.87\linewidth]{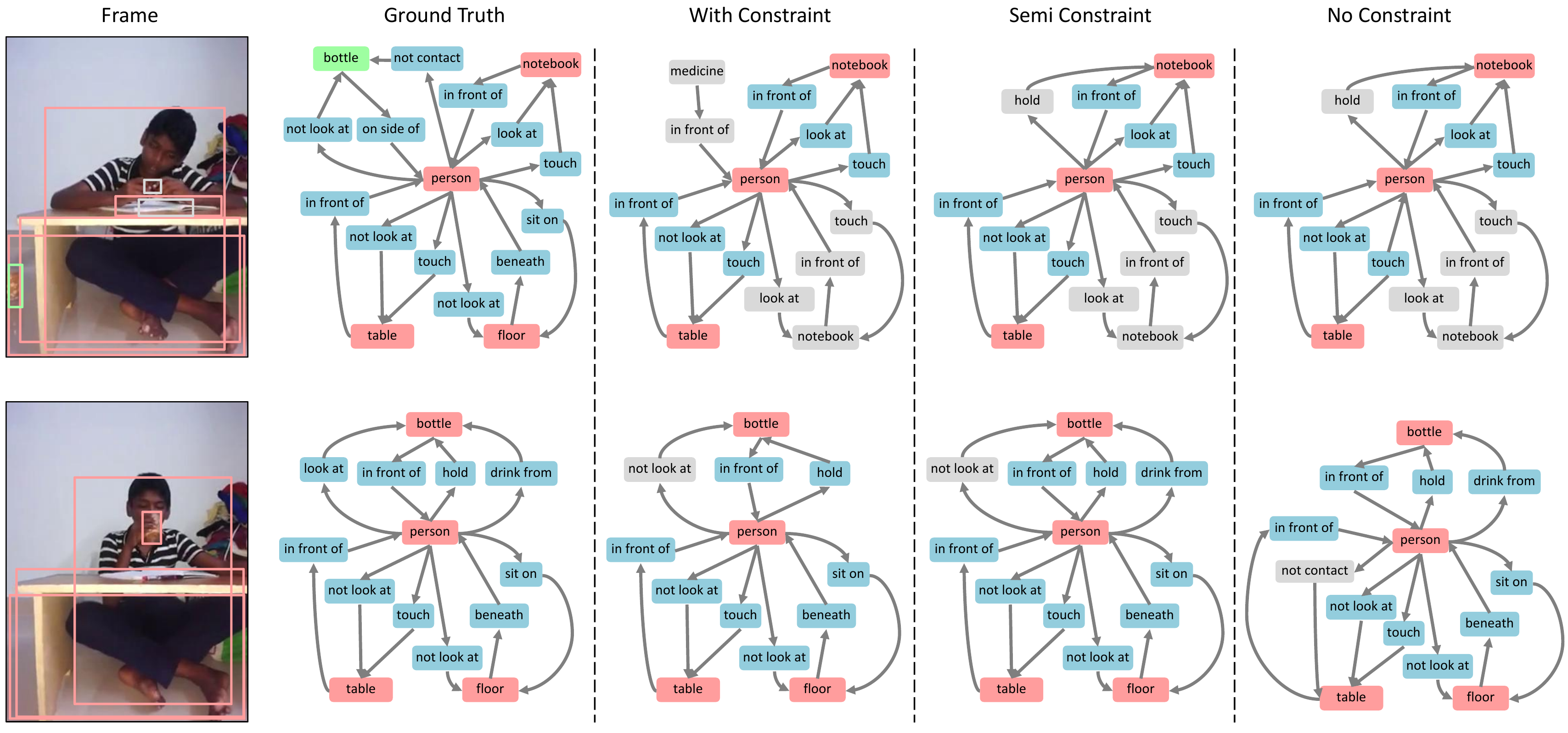}
\caption{Qualitative results for dynamic scene graph generation. The scene graphs from STTran are generated with the top-10 confident relationship predictions with different Strategies. The green box is the undetected ground truth. The melon and gray colors indicate true positive and false positive respectively. Correct relationships are colored with light blue for clarity and  $ing$ is omitted for brevity. It shows poor object detection results reduce the performance and the result of \textbf{Semi Constraint} is closer to the ground truth.}
  \vspace{-4mm}
\label{fig:qualitative_result}
\end{figure*}  

  \vspace{-4mm}
\paragraph{Can STTran really understand temporal dependencies?}
In order to verify that STTran really improves performance through temporal dependencies in the video, instead of using clearer feature representation or powerful multi-head attention module, we trained our model with the processed training set and show the results in Table \ref{tab:train_data}.

We randomly sample $1/3$ videos in the training set and shuffle/reverse them. Meanwhile, the test set remains unchanged. As shown in Table \ref{tab:train_data}, PredCLS-$R@20$ (\textbf{With Constraint}) drops significantly from 71.8\% to 71.0\%, when one-third of the training videos are reversed, which is equivalent  to adding noise in the temporal information. Moreover, shuffled videos indicate the temporal information is completely broken and the noise is further amplified. The experimental result (first row) is in line with expectations: PredCLS-$R@20$ drops to 70.6\%. The experiments demonstrate where the improvement comes from and validate that the temporal dependencies are learned in STTran.

\begin{table}[!htbp]
\centering
\begin{adjustbox}{max width=0.48\textwidth}
\begin{tabular}{cccc}
 \hline \hline
Normal Video & Processed Video & Processing &PredCLS-R@20 \\
  \hline \hline
$2/3$ & $1/3$ & shuffle &  70.6 \\
$2/3$ & $1/3$ & reverse & 71.0  \\
$1$ & - & - & \textbf{71.8}  \\
\hline \hline
\end{tabular}
\end{adjustbox}
\caption{We shuffle/reverse one-third of the videos in the training set to explore the sensitivity of the model to frame sequence. By disorganizing the temporal information via shuffling or reversing the video sequence, the performance of the model degrades accordingly as expected.} 
  \vspace{-4mm}
\label{tab:train_data}
\end{table}


\subsection{Ablation Study}
\label{Sec:ablation}
In our Spatial-Temporal Transformer, two modules are proposed, a Spatial Encoder and Temporal Decoder. Furthermore, we integrate the temporal position into the relationship representations with the frame encoding in the Temporal Decoder. In order to clarify how these modules contribute to the performance, we ablate different components and present the results in Table \ref{tab:ablation}. We adopt PredCLS-$R@20$ and SGDET-$R@20$ as the metrics with \textbf{With Constraint} and \textbf{Semi Constraint}. PredCLS shows the ability of relationship prediction intuitively while SGDET indicates the performance of scene graph generation. 

When only the spatial encoder is enabled, the model works the same as the image-based method and also has a similar performance as RelDN \cite{zhang2019graphical}. The isolated temporal decoder (second row) boosts the performance significantly with the additional information from other frames. PredCLS-$R@20$ is improved slightly when the encoder and decoder both work whereas the improvement of SGDET-$R@20$ is limited by the object detection backbone. The learned frame encoding helps STTran fully understand the temporal dependencies and has a strong, positive effect both on PredCLS-$R@20$ and SGDET-$R@20$ while the fixed sinusoidal encoding performs unsatisfactorily. Two instances respectively predicted by the spatial encoder only and the complete STTran are shown in Fig.~\ref{fig:ablation_instance}. Without temporal dependencies, the spatial encoder mistakenly predicts \texttt{$<$person-eating-food$>$} as \texttt{$<$person-touching-food$>$} in the second frame whereas STTran infers the relationship correctly. This explicitly proves that STTran can utilize temporal context to improve scene graph generation.
\begin{table}[!htbp]
\centering
\begin{adjustbox}{max width=0.48\textwidth}
\begin{tabular}{ccccccc}
 \hline \hline
Spatial &Temporal &Frame &\multicolumn{2}{c}{PredCLS-R@20} & \multicolumn{2}{c}{SGDET-R@20} \cr
 \cmidrule(lr){4-5}  \cmidrule(lr){6-7}
Encoder &Decoder &Encoding &With &Semi &With &Semi \\
  \hline \hline
\checkmark & - & - &  69.6 & 78.7 &32.9 &35.1\\
- & \checkmark & - & 71.0 & 82.2 &33.7 &35.5 \\
\checkmark & \checkmark & - & 71.3& 82.7 &33.8 &35.6  \\
\checkmark & \checkmark & sinusoidal & 71.3 &82.8 &33.9 &35.7\\
\checkmark & \checkmark & learned & \textbf{71.8} &\textbf{83.1} &\textbf{34.1} &\textbf{35.9}\\
\hline \hline
\end{tabular}
\end{adjustbox}
  \caption{Ablation Study on our STTran. $\checkmark$ indicates the corresponding module is enabled while $-$ indicates disabled. We also compare the effectiveness of sinusoidal and learned positional encoding.}
  \vspace{-4mm}
  \label{tab:ablation}
\end{table}


\subsection{Qualitative Results}
Fig.~\ref{fig:qualitative_result} shows the qualitative results for the dynamic scene graph generation. The five columns from left to right are RGB frame, scene graph generated by ground truth, scene graph generated with the top-10 confident relationship predictions with the Strategies \textbf{With Constraint}, \textbf{Semi Constraint} and \textbf{No Constraint}. The melon color indicates truth positive whereas gray indicates false positive. The green box is the ground truth not detected by the detector. In the first row, two false positives with high object detection confidence (\textsl{medicine} and \textsl{notebook}) result in wrong predictions among the top-10 relationships. All the top-10 confident relationships following three strategies are of high quality in the second row when the object detection is successful. \texttt{$<$person-drinking from-bottle$>$} in the third column is lost because \textbf{With Constraint} only allows at most one relationship between each subject-object pair for each type of relationship while \texttt{$<$person-not contacting-bottle$>$} replaces the attention relationship between $person$ and $bottle$ in the top-10 confident list when using \textbf{No Constraint}. The two frames in Fig. \ref{fig:qualitative_result} are not adjacent since the detected $person$s overlap with the ground truth IoU $<0.5$ in the frames between them.


\section{Conclusion}
In this paper, we propose Spatial-Temporal Transformer (STTran) for dynamic scene graph generation whose encoder extracts spatial context within a frame and decoder captures the temporal dependencies between frames. Distinct from single-label losses in previous works, we utilize a multi-label margin loss and introduce a new strategy to generate scene graphs. Several experiments demonstrate that temporal context has a positive effect on relationship prediction. 
We obtain state-of-the-art results for the dynamic scene graph generation task on the Action Genome dataset. 

\vspace{2mm}
\noindent
\textbf{Acknowledgements }
This work was supported by the DFG PhoenixD (EXC 2122) and COVMAP (RO 2497/12-2).

\section{Appendix}
\noindent
In this supplementary material, we provide additional implementation details for our method in Sec.~\ref{sec:implem} of this appendix.
In Sec.~\ref{sec:ag}, we present detailed analysis of the Action Genome dataset \cite{ji2020action}.
In Sec.~\ref{sec:result}, we show additional qualitative results. Failure cases of our method are shown in Sec.~\ref{sec:fail}.

\subsection{Implementation Details}
\label{sec:implem}
In this section, we present some implementation details that were omitted in the main paper for brevity.

\begin{figure}[ht!]
\centering
\includegraphics[width=0.99\linewidth]{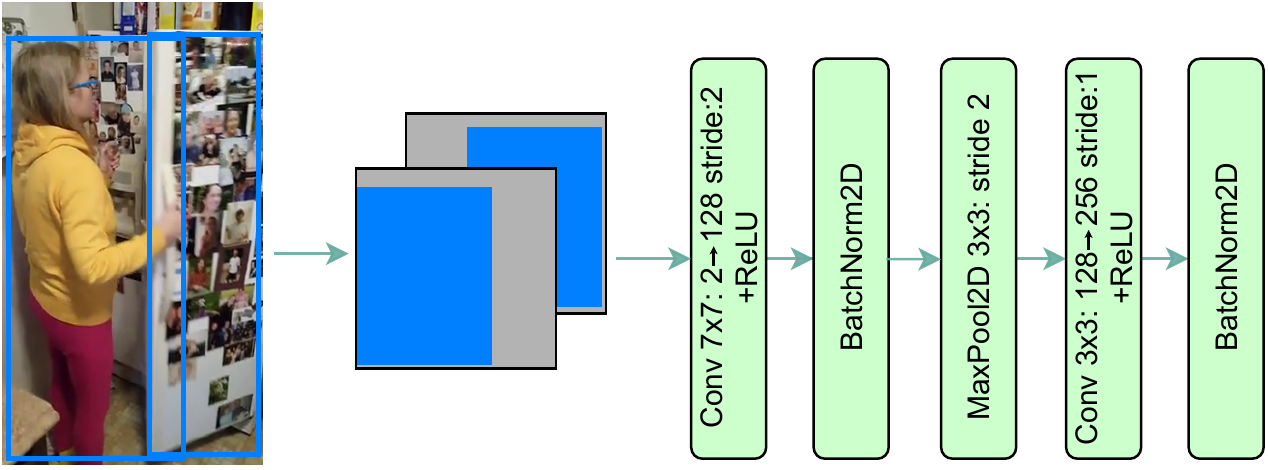}
\caption{Illustration of the box function $f_{box}$}
\label{fig:fbox}
  \vspace{-4mm}
\end{figure}

\vspace{-2mm}
\paragraph{Box Function $f_{box}$} It transforms the bounding boxes of the subject and object to the  $256\cdot7\cdot7$ feature map. 
Follow-ing \cite{zellers2018neural}, the bounding boxes of the subject and object are firstly converted to a binary spatial mask of size $2\cdot27\cdot27$ which indicates the location of the subject and object in the frame. By forwarding the spatial mask into a convolutional network (see Fig.~\ref{fig:fbox}), the location representation is computed which can be added to the $256\cdot7\cdot7$ feature map of the union box.

\vspace{-2mm}
\paragraph{Queries and Keys in the Temporal Decoder} For the $i$-th batch in the decoder layers, the queries $\bm{Q}$ and keys $\bm{K}$ are computed by adding the learned frame encoding $ \bm{E}_{f} = [\bm{e}_{1},\dots,\bm{e}_{\eta}]$ to $\bm{Z}_{i} = [\bm{X}_{i},\dots,\bm{X}_{i+\eta-1}]$. Note that $\bm{E}_{f}$ and $\bm{Z}_{i}$ have the same length. $\bm{X}_i=\lbrace \bm{x}^1_i,\dots,\bm{x}^{K(i)}_i\rbrace$ denotes all the relationship representations in the $i$-th frame. Here we use braces to emphasize that there is no order between relationships in the same frame and $\bm{X}_{i}$ is still a matrix (tensor) in our PyTorch code. Therefore, the first elements of $\bm{Q}$ and $\bm{K}$ can be formulated as:
  \begin{equation}
        \bm{q}_1 = \bm{k}_1 = \bm{e}_{1}+\bm{X}_i = [\bm{x}^1_i+\bm{e}_{1},\dots,\bm{x}^{K(i)}_i+\bm{e}_{1}]
    \label{Eq:rel_features}
    \end{equation}
which means the same encoding is added to the relation representations in the same frame.

\begin{figure}[ht!]
\centering
\includegraphics[width=0.5\linewidth]{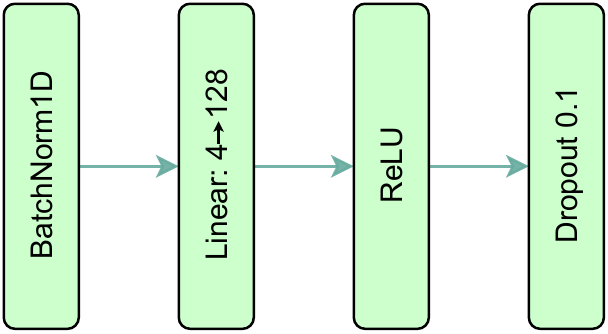}    
\caption{The box coordinate is forwarded into the feed-forward network to compute the position embedding.} 
\label{fig:object_classifier}
\vspace{-4mm}
\end{figure}

\vspace{-2mm}
\paragraph{Object Classification}
FasterRCNN~\cite{ren2015faster} based on ResNet101 outputs a $\texttt{2048-d}$ feature vector and a class distribution for each object proposal box. With multiplying the class distribution by the linear matrix $W_e\in\mathbb{R}^{36\times200}$, a $\texttt{200-d}$ semantic embedding is computed. Meanwhile, the $\texttt{4-d}$ box coordinate is forwarded into a feed-forward network (see Fig. \ref{fig:object_classifier}) to achieve a $\texttt{128-d}$ position embedding. We concatenate the feature vector, semantic embedding and position embedding, then project the concatenated vector to a $\texttt{37-d}$ distribution (including the class $background$) with two linear layers and a ReLU function in between.

\vspace{-2mm}
\paragraph{Data Pre-processing}
When performing down-sampling in the backbone, the visual information of ultra-small objects is damaged. In the experiments for SGCLS/SGDET, we only keep bounding boxes with short edges larger than 16 pixels as ~\cite{li2017scene} did. 

\begin{figure}[ht!]
\centering
\includegraphics[width=0.9\linewidth]{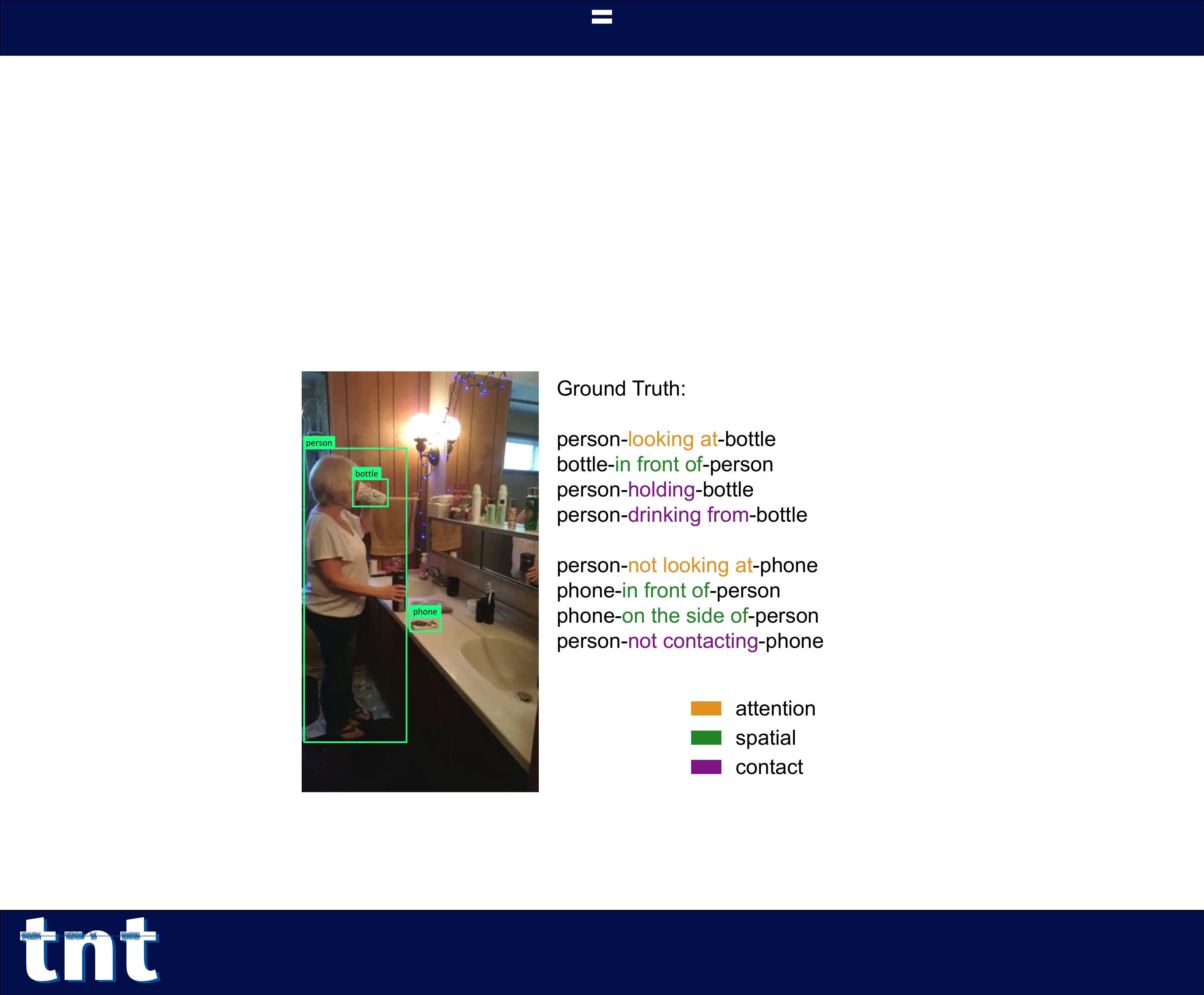}
\caption{An example of the data annotation in Action Genome dataset.} 
\vspace{-4mm}
\label{fig:data_annotation}
\end{figure}  

\subsection{Benchmark from Action Genome}
\label{sec:ag}
In the Action Genome (AG) dataset \cite{ji2020action}, each human-object pair is annotated with three types of relationships, namely \textsl{attention}, \textsl{spatial}, and \textsl{contact} relationships where \textsl{attention} and \textsl{contact} relationships are formulated in the order of \texttt{$<$person-predicate-object$>$}, and \textsl{spatial} relationships are in the order of \texttt{$<$object-
predicate-person$>$}. Note that the \textsl{spatial}, and \textsl{contact} relationships can be annotated with multiple labels in Action Genome dataset. An annotation example is shown in Fig.~\ref{fig:data_annotation}.

A benchmark following \textbf{With Constraint} is provided by \cite{ji2020action}. However, their evaluation code and object detector have not been released. We also evaluate several advanced image-based models. Although the ranking of the model performances is consistent with \cite{ji2020action} (\texttt{VRD} \cite{lu2016visual}$<$\texttt{Motif Freq} \cite{zellers2018neural}$<$\texttt{MSDN} \cite{li2017scene}$<$\texttt{RelDN} \cite{zhang2019graphical}), the values of $Recall@K$ are different. PredCLS-$R@K$ ($K=[10,20,50]$) computed by us are generally much higher, \eg, PredCLS-$R@20$ from us $=69.5$ whereas PredCLS-$R@20$ from \cite{ji2020action} $=49.4$ for RelDN \cite{zhang2019graphical}. The reason for the difference was found after discussing with the authors of \cite{ji2020action}. Each person-object pair is allowed to have either an \textsl{attention} or \textsl{contact} relationship in \cite{ji2020action}. Instead of, we allow each person-object to have:
\begin{itemize}
\item $<$person-attention relationship-object$>$
\item $<$object-spatial relationship-person$>$
\item $<$person-contact relationship-object$>$
\end{itemize}
for \textbf{With Constraint} so that $attention$ and $contact$ relationships can be detected simultaneously. Each human-object pair is allowed to have more than one $spatial$ or $contact$ relationship when the confidence score is higher than the threshold ($0.9$) following \textbf{Semi Constraint}. For \textbf{No Constraint}, the most confident top-$K$ relationships are chosen no matter what kind of relationship.
SGCLS/SGDET-$R@K$ ($K=[10,20,50]$) from \cite{ji2020action} are slightly higher than ours. We argue that their object detector has a better performance which is crucial for SGCLS/SGDET. Note that person boxes in the ground truth are annotated by the detector from \cite{ji2020action} in the present version of the Action Genome dataset.

Furthermore, there are two kinds of $Recall@K$ metrics in \cite{ji2020action}: image-wise and video-wise. The video-wise $Recall@K$ is not adopted in our work because the only difference is whether the per-frame measurements in each video are ﬁrst averaged. 

\begin{figure}[http]
\centering
\includegraphics[width=0.98\linewidth]{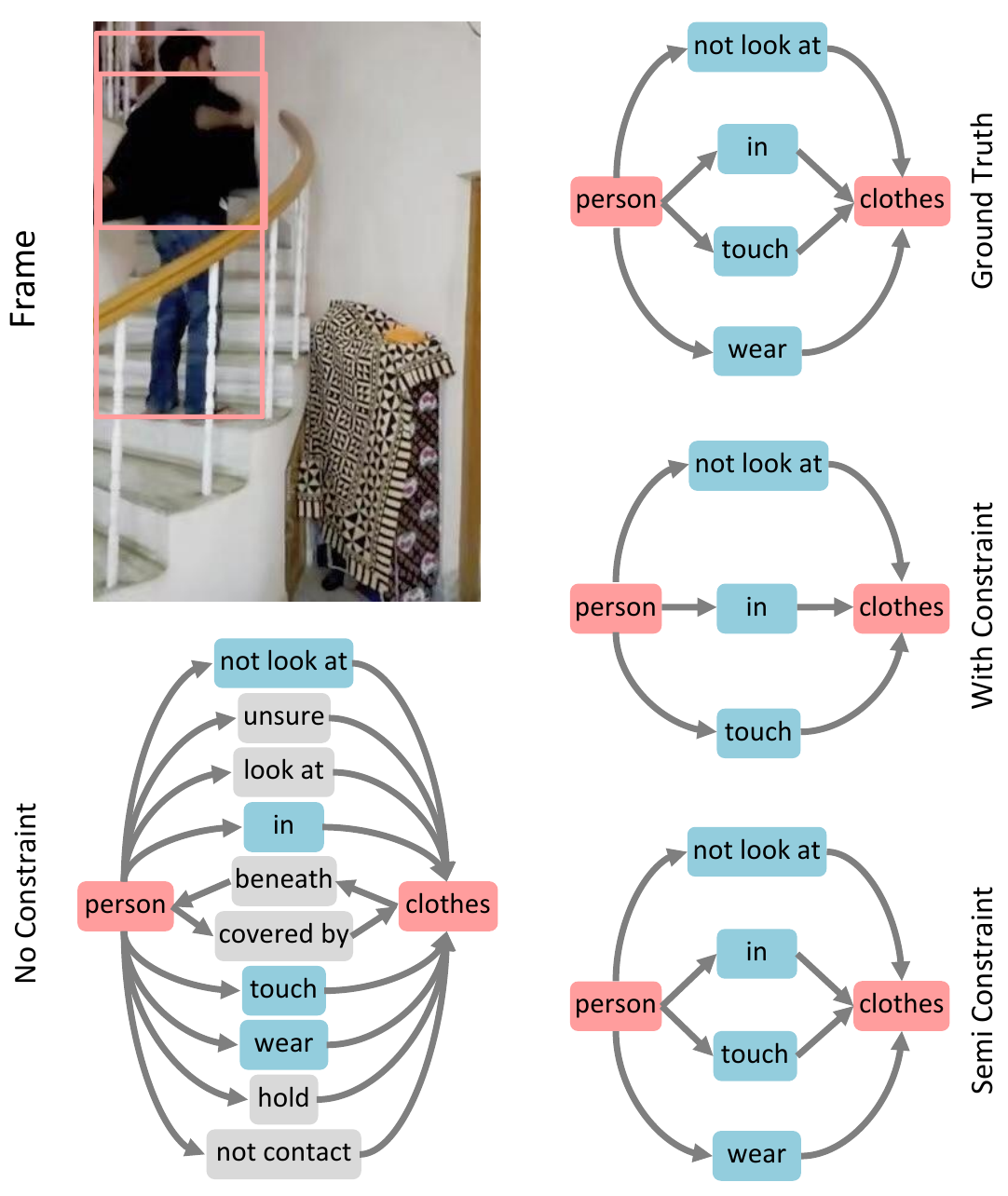}
\caption{Qualitative instances in the PredCLS setting following different strategies. Light blue relationships are true positives predicted by STTran at the $R@10$ setting while gray are false positives. 
The graph from \textbf{Semi Constraint} is identical to the ground truth, whereas there are several false positives in the graph from \textbf{No Constraint} without restriction.
} 
  \vspace{-4mm}
\label{fig:semi_constraint}
\end{figure}

\begin{table*}[ht!]
\centering
\begin{adjustbox}{max width=1\textwidth}
\begin{tabular}{cccccccccccc}
 \hline \hline
 \multirow{2}*{Method} &  \multicolumn{10}{c}{$AP_{pred}$} \cr
    \cmidrule(lr){2-12} 
&not looking at &looking at &in front of &on the side of &beneath &hebind &holding &not contacting &touching &sitting on &\textsl{mean}\\
  \hline \hline
  GPS-Net\cite{lin2020gps}& 64.94 & 49.81 & 90.14& 38.08& \textbf{88.98}& 77.45& \textbf{88.38}& \textbf{81.30}& 37.26 & 88.36 & \textsl{70.47}\\
  STTran& \textbf{79.73} & \textbf{67.07}& 90.14& \textbf{40.52}& 88.93& \textbf{81.01}& 85.29& 81.29& \textbf{37.50}& \textbf{90.67} & \textbf{\textsl{74.22}}\\
\hline \hline
\end{tabular}
\end{adjustbox}
  \caption{The average precision of predicates $AP_{pred}$ for the top-10 frequent relationships including 2 $attention$, 4 $spatial$ and 4 $contact$ relationships. We compare our model with GPS-Net \cite{lin2020gps} which performs best on the Action Genome dataset among the image-based baselines. With temporal dependencies, STTran has a great advantage in predicting $attention$ relationships and also performs better for $spatial$ relationships. For $contact$ relationships, GPS-Net outperforms STTran on the prediction of $holding$ and $not\; contacting$. The last column is the mean of $AP_{pred}$ for these 10 relationships.}
  \vspace{-1mm}
  \label{tab:map_pred}
\end{table*}

\subsection{Additional Results}
\label{sec:result}
We also report the average precision of predicates $AP_{pred}$ to evaluate the performance for single relationships. The $AP_{pred}$ evaluates the average precision of the predicates where the subject and object boxes are given. The 10 most frequently occurring relationships in Action Genome dataset (2 $attention$, 4 $spatial$ and 4 $contact$ relationships) are evaluated with our model and GPS-Net \cite{lin2020gps}, which performs best in the image-based scene graph generation methods. The results are shown in the Table~\ref{tab:map_pred}. Compared with GPS-Net, our model has a great advantage in predicting $attention$ relationships with temporal dependencies and also performs better for $spatial$ relationships. However, GPS-Net outperforms STTran on the prediction of $holding$ and $not\; contacting$ for $contact$ relationships.


Different performance of 3 generation strategies are demonstrated in Fig.~\ref{fig:semi_constraint}. 
For \textbf{With Constraint}, $wearing$ is abandoned since only one contact relationship is allowed between each object pair. Although \textbf{No Constraint} allows multi-label prediction, the result contains a lot of noise when there are few pairs in the frame, especially bounding boxes are given in PredCLS and SGCLS.

\begin{figure*}[http]
\centering
\includegraphics[width=0.9\linewidth]{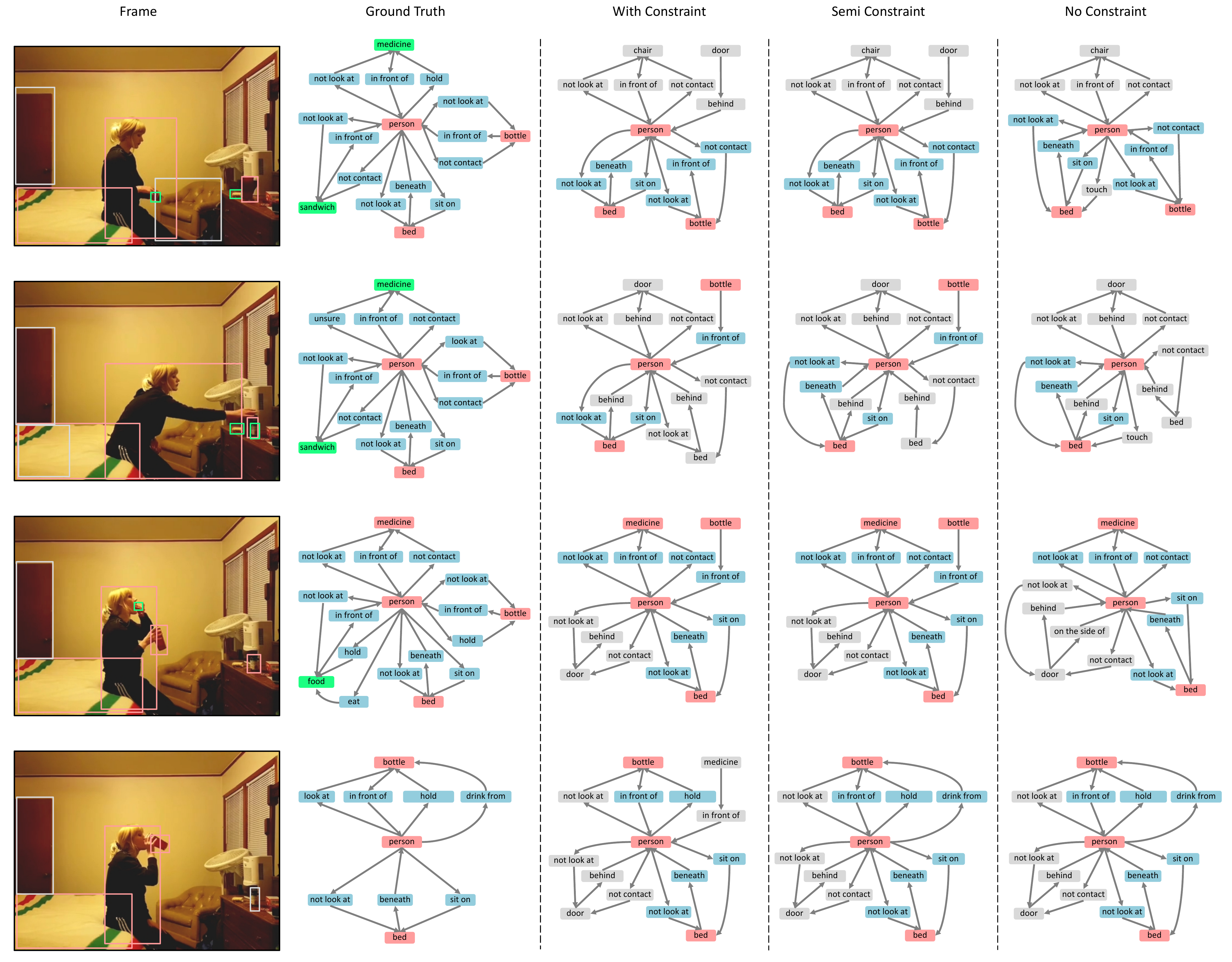}
\caption{Qualitative results for dynamic scene graph generation. The scene graphs are generated with the top-10 confident predictions with different Strategies in the SGDET task. The green boxes denote the undetected ground truth. The melon and gray colors indicate true positive and false positive respectively. Correct relationships are colored with light blue whereas relationships not in the ground truth are colored with gray. In the video the person sitting on the bed holds the medicine and bottle. Then she takes the medicine and drinks water from the bottle.}
\label{fig:add_qresults}
\end{figure*}

Additional qualitative results for dynamic scene graph generation from the video are shown in Fig.~\ref{fig:add_qresults}.  The dynamic scene graphs are generated with the top-10 confident predictions with different Strategies in the SGDET task.  The green boxes denote the undetected truths. The melon and gray colors indicate true positive and false positive respectively. Correct relationships are colored with light blue whereas relationships not in the ground truth are colored with gray. In the video the person sitting on the bed holds the medicine and bottle. Then she takes the medicine and drinks water from the bottle.

\begin{figure*}[http]
\centering
\includegraphics[width=0.6\linewidth]{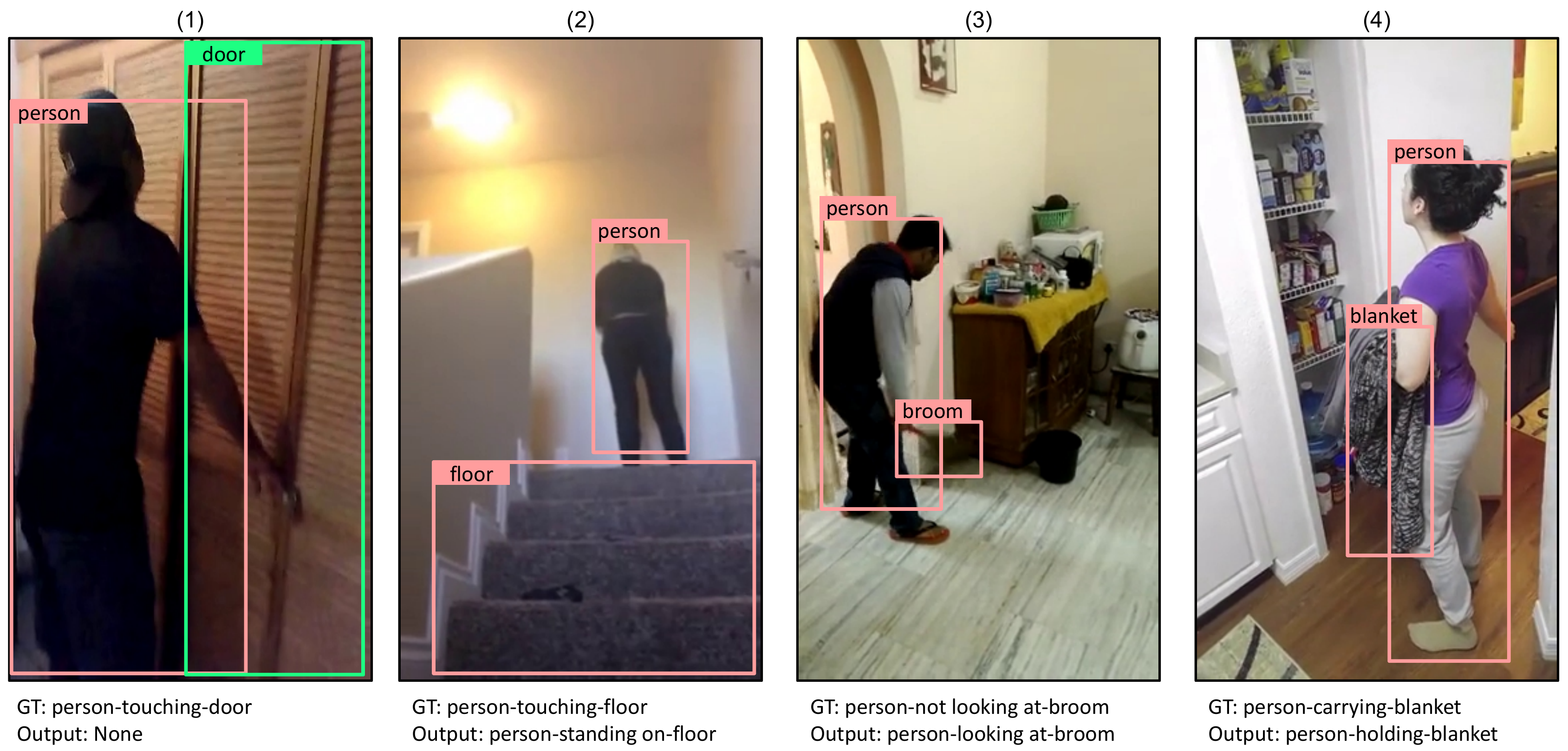}    
\caption{Instances of the most common failure cases. (1) The door is not detected by the object detector and there is no corresponding relationships in the output. (2) STTran predicts that the person is standing on the floor while the ground truth is incorrect. (3) Although the prediction from STTran is wrong, it is difficult for humans to identify whether the person is looking at the broom or not. (4) $carrying$ which occurs less frequently in Action Genome is predicted as $holding$ with a similar meaning and a higher frequency.}
\label{fig:cases}
\end{figure*}

\subsection{Failure Cases}
\label{sec:fail}
In order to clarify the limitation of the model, we analyze the results and summarize the following most  common failure cases:
\begin{itemize}
\item[1.] The object is not detected (IoU$<0.5$), particularly small objects such as $phone$ and $medicine$.
\item[2.] The predictions do not match the ground truth relationships which are annotated by mistake.
\item[3.] The relationship is ambiguous and difficult to be identified even by humans. 
\item[4.] The model predicts the wrong majority relationship instead of the correct minority relationship.
\end{itemize}
The failure cases are shown in Fig.~\ref{fig:cases}. We conjecture that Failure 1
can be improved by a better object detector. Failure 2 and Failure 3 are caused by the human-labeled annotations. Failure 4 is caused by the imbalanced relationship distribution both in the dataset and in the real world.

{\small
\bibliographystyle{ieee_fullname}
\bibliography{egbib}
}
\end{document}